\documentclass[10pt,twocolumn,letterpaper]{article}

\usepackage{wacv}              

\usepackage{graphicx}
\usepackage{amsmath}
\usepackage{amssymb}
\usepackage{booktabs}
\usepackage{amssymb}
\usepackage{amsthm}
\usepackage[figuresright]{rotating}
\usepackage{amsmath}
\usepackage{pifont}
\usepackage{mathrsfs}
\usepackage{newtxmath}
\usepackage{amsfonts}
\usepackage{algorithm}
\usepackage{algorithmic}
\usepackage{soul}
\usepackage{url}
\usepackage[utf8]{inputenc}
\usepackage{caption}
\usepackage{subcaption}
\usepackage{graphicx}
\usepackage{booktabs}
\usepackage{xcolor}
\usepackage{multirow}
\usepackage{float}
\usepackage{comment}
\usepackage{stackengine}
\usepackage{mathtools, nccmath}
\DeclareMathOperator{\softmax}{softmax}

\DeclareMathOperator*{\argmin}{arg\,min}
\newcommand{\cmark}{\ding{51}}%
\newcommand{\xmark}{\ding{55}}%

\usepackage[pagebackref,breaklinks,colorlinks]{hyperref}

\usepackage[capitalize]{cleveref}
\crefname{section}{Sec.}{Secs.}
\Crefname{section}{Section}{Sections}
\Crefname{table}{Table}{Tables}
\crefname{table}{Tab.}{Tabs.}

\def\wacvPaperID{1506} 
\def\confName{WACV}
\def\confYear{2025}

\begin{document}


\title{GlobalDoc: A Cross-Modal Vision-Language Framework for Real-World Document Image Retrieval and Classification}

\author{Souhail Bakkali{$^1$}~Sanket Biswas{$^4$}~Zuheng Ming{$^2$}~Mickaël Coustaty{$^1$}\\~Marçal Rusiñol{$^3$}~Oriol Ramos Terrades{$^4$}~Josep Lladós{$^4$}\\
{$^1$}L3i, La Rochelle Université, France\\
{$^2$}L2TI, Université Sorbonne Paris Nord, France\\
{$^3$}AllRead Machine Learning Technologies, Spain\\
{$^4$}Computer Vision Center, Universitat Autònoma de Barcelona, Spain\\
{\tt\small \{souhail.bakkali, mickael.coustaty\}@univ-lr.fr},~{\tt\small zuheng.ming@univ-paris13.fr} \\ {\tt\small marcal@allread.ai},~{\tt\small\{sbiswas, oriolrt, josep\}@cvc.uab.cat}}

\maketitle

\begin{abstract}
   Visual document understanding (VDU) has rapidly advanced with the development of powerful multi-modal language models. However, these models typically require extensive document pre-training data to learn intermediate representations and often suffer a significant performance drop in real-world online industrial settings. A primary issue is their heavy reliance on OCR engines to extract local positional information within document pages, which limits the models' ability to capture global information and hinders their generalizability, flexibility, and robustness. In this paper, we introduce \textbf{GlobalDoc}, a cross-modal transformer-based architecture pre-trained in a self-supervised manner using three novel pretext objective tasks. GlobalDoc improves the learning of richer semantic concepts by unifying language and visual representations, resulting in more transferable models. For proper evaluation, we also propose two novel document-level downstream VDU tasks, Few-Shot Document Image Classification (DIC) and Content-based Document Image Retrieval (DIR), designed to simulate industrial scenarios more closely. Extensive experimentation has been conducted to demonstrate GlobalDoc's effectiveness in practical settings.
\end{abstract}


\section{Introduction}
\label{sec:intro}
\begin{figure}[t]
\centering
\begin{subfigure}{.5\textwidth}
  \includegraphics[width=\linewidth]{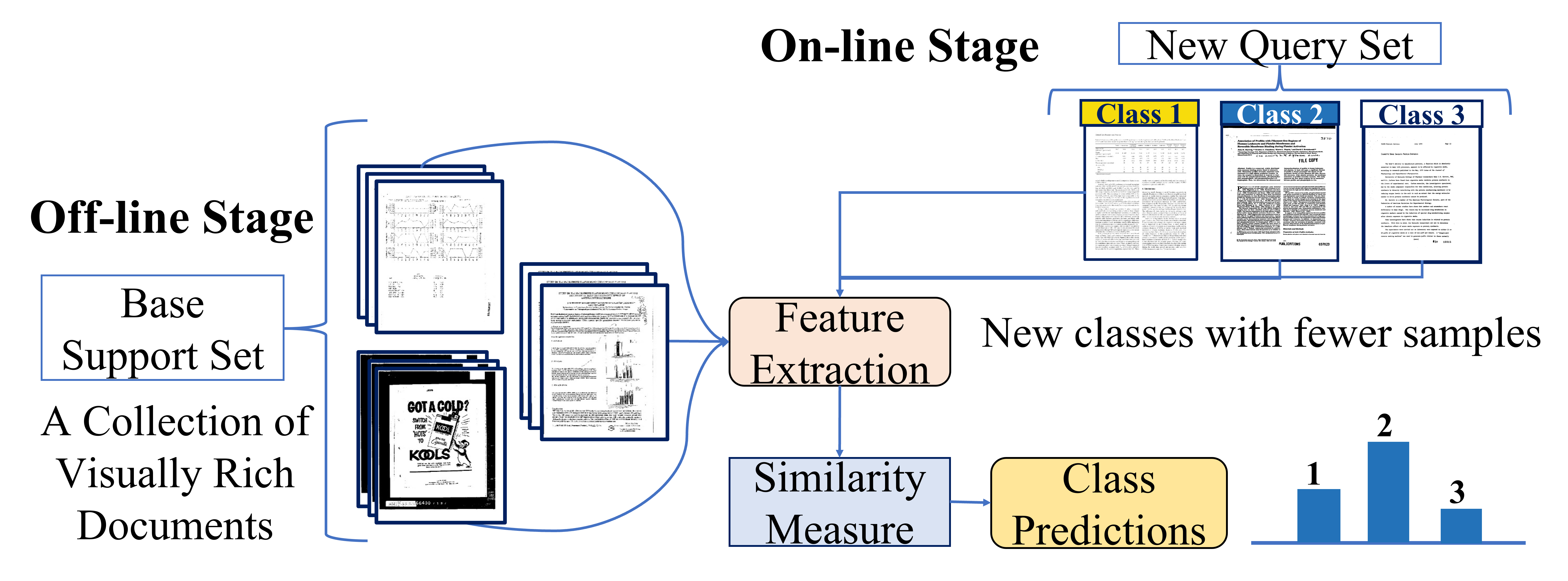} \quad
  \caption{Overview of the Few-Shot Document Classification process}
  \label{fig:fewshot_document_classification}
\end{subfigure}\hfill
\begin{subfigure}{.5\textwidth}
  \includegraphics[width=\linewidth]{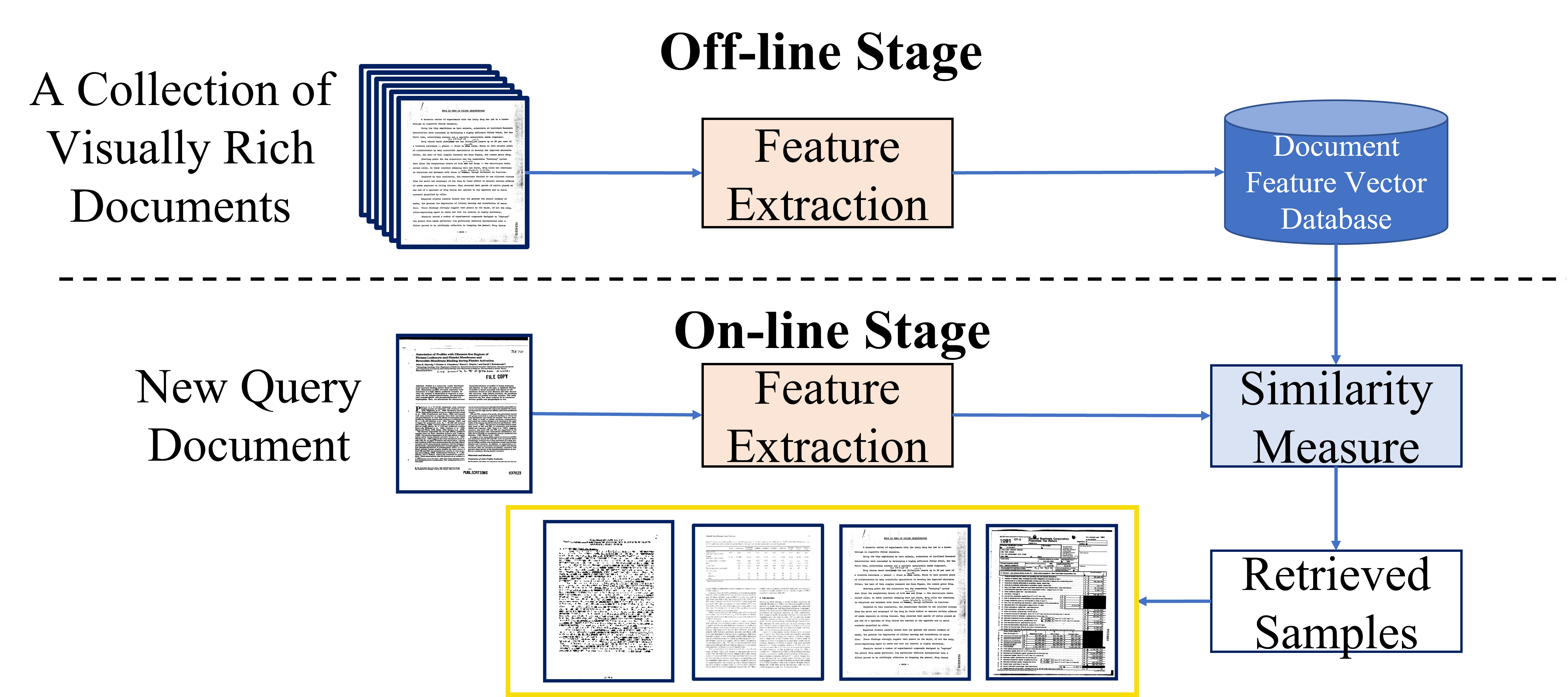} \quad
  \caption{Overview of the Content-based Document Retrieval process}
  \label{fig:document_retrieval}
\end{subfigure}
\caption{\textbf{Novel Document-level Downstream Tasks:} We propose a new modification to the "pretrain-then-finetune" paradigm for a "closer-to-real" industrial scenario with offline and online stages.}
\label{fig:industrial_data_paradox}
\end{figure}


Document AI~\cite{cui2021document} strives towards automating and accelerating the process of reading, understanding, classifying and extracting information from documents which can be of any kind (\eg digitally born, scanned and/or web-pages). With the increasing traffic in information management workflows for different industrial sectors (finTech, legalTech, insuranceTech), there is an essential need to evaluate the SOTA VDU models in an online setting, which could simulate a ``closer-to-real'' industrial scenario. The current self-supervised SOTA VDU frameworks~\cite{appalaraju2021docformer, Unifieddoc, huang2022layoutlmv3, li2022dit, li2021selfdoc, powalski2021going, xu2020layoutlmv2} have exhibited superior performance in benchmark tasks for document image understanding, such as DIC~\cite{harley2015evaluation}, visual information extraction (VIE)~\cite{jaume2019funsd} and document visual question answering (DocVQA)~\cite{mathew2021docvqa}. 

However, when it comes to deploying such models in real-world business cases in the Document AI industry, there is a significant drop in model performance and efficiency. The primary causes that contribute to this issue are: (1) \textbf{Dependency on Commercial OCR Engines:} Most of the self-supervised learning-based (SSL) approaches like DocFormer~\cite{appalaraju2021docformer} and the LayoutLM family~\cite{xu2020layoutlm,xu2020layoutlmv2,huang2022layoutlmv3} rely on pre-training with millions of document samples and heavily depend on top commercial OCR engines to learn local spatial information (\eg words along with their 2D and/or 1D positional encoding) to boost their performance for downstream VDU tasks; (2) \textbf{High Computational Demand:} Other SSL-based approaches, like SelfDoc~\cite{li2021selfdoc} and UDoc~\cite{Unifieddoc}, use fewer pre-training samples but rely on visual features at a region level (\eg, cropped semantic object regions like tables and figures) with pre-trained object detectors in their cross-modal encoded page representation. This reliance results in high computational demand; (3) \textbf{Unrepresentative Training Data:} The offline document collections used for pre-training these models are not sufficiently representative of the online documents from new categories, leading to performance issues when encountering new data; (4) \textbf{Flexibility in Multi-modal Processing:} Multi-modal VDU models require vision-language sample pairs during the online stage to perform downstream tasks. Efficiently handling large volumes of new incoming documents requires the ability to transfer knowledge from pre-analyzed offline sources. The models need to be adaptable enough to process both uni-modal (vision-only or language-only) and multi-modal document images, optimizing memory usage and computational efficiency.

Therefore, instead of limiting VDU frameworks to learn local features at the word or sentence level with their relative 2D and/or 1D positional encodings, we introduce a \textit{page-level pre-training framework} which learns the global information and the overall relationships between vision and language modalities guided by the model itself, similar to~\cite{radford2021learning,tan2019lxmert}. Our model uses a simple vision transformer (ViT) backbone~\cite{dosovitskiy2020image} for visual feature extraction, eliminating the need for a dense object detector. To further address the challenges of industrial online settings, we introduce two novel document-level downstream tasks for the VDU domain. The first task, \textit{Few-Shot DIC}, aims to adapt the pre-trained embedding model to "task" changes, simulating a "closer-to-real" industrial setting, particularly during the online phase. The second task, \textit{Content-based DIR}, evaluates our model's representation learning capability in both uni-modal and cross-modal document image retrieval settings. This task integrates language and vision cues to create a more versatile and flexible system for industrial document data usage, as illustrated in ~\cref{fig:industrial_data_paradox}. 
The key contributions of our work can be summarized in two-folds: 
\begin{itemize}
  \item We propose three novel pretext tasks for our page-level document pretraining strategy: two cross-modal objectives, \textit{learning-to-mine (L2M)} and \textit{learning-to-unify (L2U)}; and a \textit{learning-to-reorganize (L2R)} task to re-structure the representation space for each modality. 
  \item We developed two new downstream tasks to better evaluate GlobalDoc's performance in challenging industrial online settings, and showcase it's generalizing capabilities and assessment in both uni-modal and cross-modal scenarios.  
  \item We conduct a comprehensive analysis and evaluation of our model's performance on standard VDU benchmark tasks (DIC and VIE), to show the effectiveness of vision-and-language based approaches when compared with SOTA methods utilizing the relative positional encodings (as layout modality).  
\end{itemize}

\section{Related Work}
\label{sec:related}
\subsection{Self-Supervised Pre-Training}
Self-supervised methods~\cite{chen2020simple, misra2020self} in the computer vision literature primarily focus on designing suitable pretext tasks to leverage high-level semantic and compact representations from a large-scaled unlabeled corpus during the pre-training phase. These representations are then later utilized in solving downstream application tasks like classification~\cite{dosovitskiy2020image, touvron2021training} and object detection~\cite{carion2020end, zhu2020deformable}. In this context, pre-trained natural language processing (NLP) models like BERT~\cite{devlin2018bert} and RoBERTa~\cite{liu2019roberta} have also shown immense potential in generating contextualized representations from unlabeled text corpus by using masked language modelling (MLM) and next sentence prediction (NSP) pretext tasks. Vision models later incorporated a BERT-like pre-training inspired approach for images~\cite{bao2021beit, he2022masked} to capture the relationship between patches to achieve state-of-the-art performance in Self-supervised learning-based image recognition benchmarks. Alternatively, contrastive-based SSL techniques~\cite{chen2020simple, he2020momentum} were also used extensively to learn a global metric space from unlabeled image samples, which could benefit training large models without overfitting. However, such approaches could suffer from a loss in generalizability during pre-training when the augmented image batches have similar statistics. To overcome this drawback, Dwibedi \etal~\cite{dwibedi2021little} proposed to sample the nearest neighbors from the dataset in the learnt latent space and treat them as positives. This introduces more semantic variations in the representation space by providing a support set of embeddings and preventing an over-reliance on data augmentation for such approaches. Our work has drawn inspiration from this pre-training objective to be applied in our SSL-based document understanding framework. More recently,~\cite{estepa2023all4one} has improved over~\cite{dwibedi2021little}  by reducing the distance between neighbour representations using ”centroids” created through a self-attention mechanism. 

\subsection{Cross-Modal Document Understanding}

The success of BERT~\cite{devlin2018bert} in achieving SOTA results for various NLP tasks (sequence classification, named entity recognition, question answering etc.) has influenced vision-language research, leading to the development of models like ViLBERT~\cite{lu2019vilbert} and LXMERT~\cite{tan2019lxmert}, which learn cross-modal encoder representations using both visual and textual modalities. Approaches such as CLIP~\cite{radford2021learning} and ALBEF~\cite{li2021align} use transformer-based multimodal encoders to model visual (region-based image features) and textual (paired captions) tokens jointly. These tokens are aligned using a pairwise contrastive objective function before being fused with a cross-attention mechanism, enhancing vision and language representation learning. Current SOTA methods in multi-modal document pre-training~\cite{appalaraju2021docformer,appalaraju2024docformerv2, xu2020layoutlm, xu2020layoutlmv2, huang2022layoutlmv3, wang2022lilt} integrate 2D positional embeddings from page-level OCR with visual and textual information in a multimodal transformer framework. This induces relative spatial bias, improving local representation learning on large-scale pre-training datasets. Alternate approaches, such as SelfDoc~\cite{li2021selfdoc} and UDoc~\cite{Unifieddoc} uses region proposals of semantic object layouts (e.g., tables, figures, text regions) and OCR to learn global page representations with less pre-training data. Our work follows a similar motive in our proposed methodology. 
We introduce a cross-modal contrastive learning objective, inspired by~\cite{radford2021learning, li2021align}, to better align vision and language representations using the RVL-CDIP industry document database~\cite{harley2015evaluation}. Our approach employs a multimodal cross-attention encoder, enabling more robust and generalized representation learning for VDU systems. Previous attempts to leverage cross-modal document representation, such as learning joint embedding spaces for information extraction~\cite{zhang2020trie} and classification~\cite{bakkali2023vlcdoc, Dauphinee2019Modular, bakkali2020cross, bakkali2020visual}, were developed in a supervised setting. Our work focuses on vision-language feature learning during pre-training for unlabeled document images, followed by fine-tuning for DIC in both standard~\cite{harley2015evaluation} and few-shot learning~\cite{snell2017prototypical} scenarios. We also introduce a content-based DIR setting to further demonstrate the flexibility of our framework, similar to evaluation protocol followed in~\cite{gordo2013large, harley2015evaluation}. Additionally, we also draw inspiration from multi-page DIC benchmark tasks proposed in~\cite{van2024beyond}, which utilize the RVL-CDIP benchmark~\cite{harley2015evaluation}.

\section{GlobalDoc}
\label{sec:method}
\begin{figure*}[t]
  \centering
  \includegraphics[width=\linewidth]{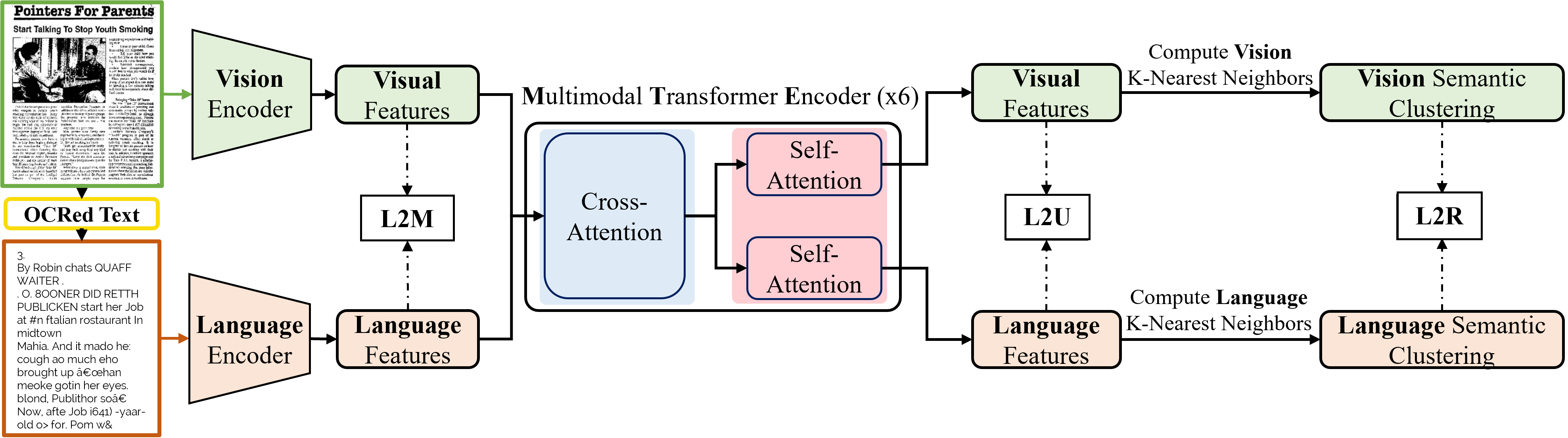}
  \caption{Overview of the proposed GlobalDoc framework with the designed pretext learning objectives.}
  \label{fig:figure_5.1}
\end{figure*}

\subsection{Model Architecture}

\noindent\textbf{Vision Encoder.} Let $v_{visn} \in \mathbb{R}^{H \times W \times C}$ be the document image which we feed to a ViT-B/16 transformer encoder~\cite{dosovitskiy2020image}. We reshape it into a sequence of flattened $2D$ patches $i_{{visn}_{p}} \in \mathbb{R}^{N \times (P^{2}\cdot C)}$, where $(H, W)=(224, 224)$ is the resolution of the document image, $C=3$ is the number of channels, $(P, P)$ is the resolution of each document patch, and $N=HW/P^2$ is the resulting number of patches, which serve as the input sequence length for the transformer encoder. The transformer encoder expects a flattened sequence as input of $d_{visn}=768$ dimension as the hidden embedding size. The resulting visual embeddings are then represented as ${V} = v^{i}_{visn} \in \mathbb{R}^{d_{visn}}$.

\noindent\textbf{Language Encoder.} Let $t_{lang}$ be the OCRed text within a document image extracted with EasyOCR\footnote{\url{https://github.com/JaidedAI/EasyOCR}}. The input sequences are fed into the pre-trained RoBERTa$_{Base}$  encoder~\cite{liu2019roberta}. The textual embeddings are denoted as \(T = t^{i}_{lang} \in \mathbb{R}^{d_{lang}}\). Next, we tokenize the plain text \(t_{lang}\) using the RoBERTa\(_{Base}\) tokenizer to obtain \(t_{tok}\). Each input sequence begins with a \([CLS]\) token and ends with a \([SEP]\) token. The tokenized sequence \(t_{tok}\) is represented as:
\(t_{tok} = [CLS], t_{{tok}_{1}}, t_{{tok}_{2}}, \ldots, t_{{tok}_{n}}, [SEP]
\)
where \(n = 256\). Sequences longer than 256 tokens are truncated, and those shorter are padded with \([PAD]\) tokens.

\noindent\textbf{Cross-Modal Attention Encoder.} The cross-modal attention encoder (CMAE) module captures intrinsic patterns by modeling both the inter-modality and the intra-modality relationships for image regions and text sequences. Specifically, the CMAE module is a transformer-based architecture as in~\cite{vaswani2017attention}, following the same design of a basic module in BERT~\cite{devlin2018bert}, with the parameters of vision-language modalities shared. It consists of a multi-head self-attention sub-layer, and a position-wise feed-forward sub-layer $f_{FF}$. Meanwhile, residual connections followed by the layer normalization $f_{LN}$ are also applied around each of the two sub-layers. In the multi-head self-attention sub-layer, the attention is calculated $h$ times, making it to be multi-headed. This is achieved by projecting the queries $\mathcal{Q}$, keys $\mathcal{K}$, and values $\mathcal{V}$ $h$ times by using different learnable linear projections. Let ${V}^l = \{v_{1}, v_{2}, ..., v_{m}\}$, ${T}^l = \{t_{1}, t_{2}, ..., t_{m}\}$ be the sets of intermediate visual and textual features at the $l$-th layer of the vision and language modalities respectively, where $v_{i}, t_{i} \in \mathbb{R}^{1 \times d_{f}}$, and ${V}, {T} \in \mathbb{R}^{m \times d_{f}}$. To accomplish the cross-modal interactions between the vision modality ${V}$ and the language modality ${T}$, the cross-attention functions are then defined as:
\useshortskip
\begin{align}
    \displaystyle 
        \text{CrossAtt}_{{V} \to {T}}({T}^{l}) &= \softmax\left(\frac{\mathcal{Q}_{{T}^l}\mathcal{K}_{{V}^l}^\top}{\sqrt{d_{k}}}\right)\mathcal{V}_{{V}^l} \\
    \displaystyle 
        \text{CrossAtt}_{{T} \to {V}}({I}^{l}) &= \softmax\left(\frac{\mathcal{Q}_{{V}^l}\mathcal{K}_{{T}^l}^\top}{\sqrt{d_{k}}}\right)\mathcal{V}_{{T}^l}
\label{eq:equation_12}
\end{align}
This way, we emphasize the agreement between the visual regions and the semantic meaning of text sequences. Further, the outputs of each vision-language cross-attention module are subsequently fed into the vision-language self-attention module to intensify the inner-modality information by establishing inner-interactions and exploring similar patterns as in ~\cite{li2021selfdoc}. We then have:

\useshortskip
\begin{align}
    \displaystyle 
    \text{SelfAtt}_{{V} \to {V}}({I}^{l}) &= \softmax\left(\frac{\mathcal{Q}_{{V}^l}\mathcal{K}_{{V}^l}^\top}{\sqrt{d_{k}}}\right)\mathcal{V}_{{V}^l}
\label{eq:equation_3}
\end{align}
Similarly, the $\text{SelfAtt}_{{T} \to {T}}({T}^{l})$ can be formulated by exchanging ${V}$ with ${T}$ in ~\cref{eq:equation_3}.

\subsection{Pre-training Objectives}
\textcolor{black}{We propose a two-step approach for self-supervised representation learning, differing from recent VDU methods. The first step introduces two new pre-training objectives: L2M and L2U. L2M aims to enhance the richness of latent representations by using nearest-neighbors to create diverse positive pairs, requiring a representative support set of embeddings. This objective helps align image and text features, facilitating cross-modal learning and improving understanding of semantic content. It also learns a shared low-dimensional space for both modalities, aiding L2U in matching more informative sample pairs. In the second step, we mine nearest neighbors based on feature similarity and integrate these semantically meaningful neighbors into a learnable framework. We then use L2R to encourage consistent and discriminative predictions for each image and text alongside their nearest neighbors.}

\noindent\textbf{Objective I: Learning-to-Mine (L2M).} The proposed learning objective (L2M) aims to force samples from language and vision that are semantically related to be closer according to the computed nearest-neighbors of each modality. As in SimCLR~\cite{chen2020simple}, a projection head is implemented on top of the vision and language backbones to map the visual-textual representations into a vector representation so that the two training schemes do not interfere with each other. The projection head is implemented as a non-linear multiple-layer perceptron (MLP) with one hidden layer, as it is more suitable for contrastive learning~\cite{chen2020simple}. Then, $L_{2}$ normalization is applied to the visual-textual embeddings so that the inner product between features can be used as distance measurements. We encourage cross-modality learning by introducing two interactive terms in the inter-modal objective defined as:
\useshortskip
\begin{align}
    \begin{aligned}
        \mathcal{L}_{Inter}^{\text{L2M}} &= \frac{1}{M} \sum_{i=1}^M - \log \frac{\exp(\langle \text{NN}(v_i, \mathcal{V}) , t_i^{+} \rangle/ \tau)} {\sum_{k=1}^{M} \exp(\langle v_i , t_k^{+} \rangle/ \tau)} \\
        &+ \frac{1}{M} \sum_{i=1}^M - \log \frac{\exp(\langle \text{NN}(t_i, \mathcal{T}) , v_i^{+} \rangle/ \tau)} {\sum_{k=1}^{M} \exp(\langle t_i , v_k^{+} \rangle / \tau)} \\
    \end{aligned}
\label{eq:equation_5}
\end{align}
where $\langle\cdot,\cdot\rangle$ computes similarity scores between sample pairs, 
$\tau$ is a scalar temperature hyper-parameter, and M is the mini-batch size. The first term in ~\cref{eq:equation_5} computes the similarity score between the nearest neighbors of the given document image $\text{NN}(v_i, \mathcal{V})$ with its corresponding text sample ${t}_{i}^{+}$. Similarly, the second term computes the similarity score between the nearest neighbors of the given text sample $\text{NN}(t_i, \mathcal{T})$ and its corresponding visual sample pair ${v}_{i}^{+}$.
\text{NN}($v_{i}, \mathcal{V}$) and \text{NN}($t_{i}, \mathcal{T}$) denote the nearest neighbour vision and language operators defined respectively as:
\useshortskip
\begin{align}
    \displaystyle
        \text{NN}(v_{i}, \mathcal{V}) = \argmin_{q_{i} \in I} \mid\mid v_{i} - q_{i} \mid\mid_{2} \\
    \displaystyle
        \text{NN}(t_{i}, \mathcal{T}) = \argmin_{q_{t} \in T} \mid\mid t_{i} - q_{t} \mid\mid_{2}
\label{eq:equation_67}
\end{align}
Similarly, the intra-modal objective is computed as:
\useshortskip
\begin{align}
    \begin{aligned}
        \mathcal{L}_{Intra}^{\text{L2M}} &= \frac{1}{M} \sum_{i=1}^M - \log \frac{\exp(\langle \text{NN}(v_i, \mathcal{V}) , v_i^{+}\rangle / \tau)} {\sum_{k=1}^{M} \exp(\langle v_i ,v_k^{+} \rangle/ \tau)} \\
        &+ \frac{1}{M} \sum_{i=1}^M - \log \frac{\exp(\langle\text{NN}(t_i, \mathcal{T}) , t_i^{+}\rangle / \tau)} {\sum_{k=1}^{M} \exp(\langle t_i , t_k^{+} \rangle / \tau)}
    \end{aligned}
\label{eq:equation_4}
\end{align}
\noindent\textbf{Objective II: Learning-to-Unify (L2U)}. The principle of this objective is to predict whether a pair of a document image and its corresponding text is unified (positive) or not unified (negative). We compute the pairwise dot-product similarity between each language sequence $t_{i}$ and document image $v_{i}$ in the mini-batch as the predictions. The target similarity between the language sequence $t_{i}$ and the document image $v_{i}$ is computed as the average of the dot-product similarity between $t_{i}$ and $t_{j}$ and the dot-product similarity between $v_{i}$ and $v_{j}$. Then, the cross-entropy loss function is computed between the targets and the predictions. Given a mini-batch with $M$ pairs, for each document image $v_{i}$, the vision-language pairs are constructed as $\left\{(v_{i}, t_{j}), y_{i,j}\right\}_{j=1}^M$, where $y_{i,j}=1$ means that $(v_{i}, t_{j})$ is a unified pair, while $y_{i,j}=0$ indicates a non-unified pair:
\useshortskip
\begin{align}
    \mathcal{P}(t_i,v_j) = \frac {\exp(\langle t_{i}, v_{j} \rangle)} {\sum^M_{k=1} \exp(\langle t_{i}, v_{k} \rangle )}
\label{eq:equation_3.6}
\end{align}
where $\mathcal{P}(t_i, v_j)$ is the probability of matching $t_{i}$ to $v_{j}$. In the L2U scenario, the unifying loss is usually computed in two directions as in~\cite{chen2019closer, wang2018learning, liu2017learning}. The $visn \to lang$ loss requires the unified text to be closer to the document image than non-unified ones, and vice versa, the $lang \to visn$ unify loss constrains the related text to rank before unrelated ones. Therefore, the L2U loss can be written as:
\useshortskip
\begin{align}
    \mathcal{L}^{\text{L2U}} = - \frac{1}{M} \sum_{i=1}^M \left (\log \mathcal{P}(v_i, t_j) + \log \mathcal{P}(t_i, v_j)\right)
\label{eq:equation_9}
\end{align}

\noindent\textbf{Objective III: Learning-to-reorganize (L2R)}.
This objective aims to leverage the pretext features learned from the L2M and L2U objectives as a prior for clustering both document images and their corresponding text sequences. We propose that a pretext task from representation learning can be used to obtain richer semantic links between vision and language modalities~\cite{van2020scan}. Specifically, for every document image $v_i \in M$ and its corresponding text sequences $t_i \in M$, we mine their $K$ nearest neighbors in the embedding space $\Phi_{\phi}$. Let $\mathcal{N}v_i$ and $\mathcal{N}t_i$ be the sets of neighboring samples of $v_i$ and $t_i$ in the mini-batch $M$, respectively. We aim to learn a clustering function $\Phi_{\eta}$ with weights $\eta$ that classifies a sample document image $v_i$ and a sample text sequence $t_i$ along with their mined neighbors $\mathcal{N}v_i$ and $\mathcal{N}t_i$ together. The function $\Phi_{\eta}$ terminates in a softmax function to perform a soft assignment over the vision clusters $\mathcal{C}_{Visn} = \{1, ..., c_{visn}\}$ and language clusters $\mathcal{C}_{Lang} = \{1, ..., c_{lang}\}$, with $\Phi_{\eta}(v_i) \in [0,1]^{C_{Visn}}$ and $\Phi_{\eta}(t_i) \in [0,1]^{C_{Lang}}$. The probabilities of sample pairs $v_i$ and $t_i$ being assigned to clusters $C_{Visn}$ and $C_{Lang}$ are denoted as $\Phi_{\eta}^{c_{visn}}(v_i)$ and $\Phi_{\eta}^{c_{lang}}(t_i)$, respectively. We then learn the weights of $\Phi_{\eta}$ by minimizing the following objective for the vision modality:
\useshortskip
\begin{equation}
\begin{aligned}
    \mathcal{L}_{Visn}^{\text{L2R}} &= -\frac{1}{|M|} \sum_{i \in M} \sum_{k \in \text{NN}(v_i, \mathcal{I})} \log \langle \Phi_{\eta}(i), \Phi_{\eta}(k) \rangle \\
    &+ \lambda \sum_{c_{Visn} \in \mathcal{C}_{Visn}} \Phi_{\eta}^{'c_{Visn}} \log \Phi_{\eta}^{'c_{Visn}} \\
    & \text{where} \quad \Phi_{\eta}^{'c_{Visn}} = \frac{1}{|M|} \sum_{i \in M} \Phi_{\eta}^{c_{Visn}}(i)
    \label{eq:equation_10}
\end{aligned}
\end{equation}
The first term in ~\cref{eq:equation_10} forces $\Phi_{\eta}$ to ensure that neighbors have the same clustering assignment, making consistent predictions for a sample document image $v_i$ and its neighboring samples $\mathcal{N}v_i$. To avoid $\Phi_{\eta}$ from assigning all document samples to a single cluster, we include the second term in ~\cref{eq:equation_10}, which is an entropy loss assigned to the clusters to ensure that the cluster distribution $\mathcal{C}$ is roughly uniform. Similarly, the language modality loss $\mathcal{L}_{Lang}^{\text{L2R}}$ is computed in the same manner. In general, the number of clusters is unknown. However, similar to prior works~\cite{li2021selfdoc}, we choose $\mathcal{C}_{Visn}$ and $\mathcal{C}_{Lang}$ equal to the number of ground-truth categories for evaluation purposes.

Hence, the combined pre-training objective to minimize $\mathcal{L}_{\text{total}}$ is given by:
\useshortskip
\begin{equation}
    \mathcal{L}_{\text{total}} = \mathcal{L}^{\text{L2M}} + \mathcal{L}^{\text{L2U}} + \mathcal{L}^{\text{L2R}}
\end{equation}
\subsection{Novel Document-level Downstream Tasks}
\noindent\textbf{Task I: Few-Shot Document Image Classification.}
To conduct the few-shot document image classification task, we use the pre-trained embedding network from stage one (\ie pre-training), and then apply meta-learning with an episodic manner. The task is illustrated as a K-way C-shot problem, as illustrated in~\cref{fig:fewshot_document_classification}. Given C labelled samples for each unseen class, the model should fast adapt to them to classify novel classes. The entire test set can be presented by $D = \left\{ [(v_{1}, y_{1}), \ldots, (v_{N}, y_{N})], [(t_{1}, y_{1}), \ldots, (t_{N}, y_{N})] \right\}$ where $N$ is the total number of classes in $D$, and $v,t$ are samples from the test set with label $y$. For a specific K-way C-shot meta-task $T$, $Y = \left\{yi|i = 1, . . . ,K\right\}$ denotes class labels randomly chosen from dataset $D$. Samples from these classes are randomly chosen to form a Support set and a Query set: (a) the support set for task $T$ is denoted as $S$, which contains C $\times$ K samples (\ie K-way $\times$ C-shot); (b) the query set is $Q$ where $n$ is the number of samples selected for meta-learning. During the meta-learning stage, the proposed model is trained to learn an embedding function to map all input images and text samples from the same class to a mean vector $c$ in a description space as a class descriptor for each class. For class $k$, it is represented by the centroid of embedding features of test samples, obtained by:
\useshortskip
\begin{align}
    \mathcal{C}_{k}= \frac{1}{|S_{k}|} \sum_{(v_{i}, t_{i}, y_{i}) \in S}\mathcal{F}(v_{i}, t_{i})
\label{eq:equation_11}
\end{align}
where $F(i_{i}, t_{i})$ is the embedding function initialized by the pretext task, $S_{k}$ are the test samples labelled with class $k$. As a metric learning-based method, we employ a distance function $d$ and produce a distribution over all classes given a query sample $q$ from the query set $Q$:
\useshortskip
\begin{align}
    \mathcal{P}{(y=k|q)} = \frac{\exp{(-d(f(q), c_{k}))}}{\sum^{K}_{k'} \exp(-d(f(q), c_{k'}))}
    \label{eq:equation_4.3.2.2}
\end{align}
Euclidean distance is chosen as distance function $d$. As shown in ~\cref{eq:equation_4.3.2.2}, the distribution is based on a softmax over the distance between the embedding of the samples (in the query set) and the class descriptors. The loss in the meta-learning stage can then read:
\useshortskip
\begin{align}
    \mathcal{L}_{meta} = d(f(q), c_{k}) + \log {\sum_{k'}d(f(q), c_{k'})}
    \label{eq:equation_4.3.2.3}
\end{align}
\noindent\textbf{Task II: Content-based Document Image Retrieval.}
To conduct the Content-based Document Retrieval task, we intend to evaluate the generalizability of GlobalDoc on both uni-modal and cross-modal retrieval tasks on each setting. We would like to answer the research question of \textit{the utility of cross-modal representations encoded by the pre-trained GlobalDoc to solve query tasks in both uni-modal and cross-modal retrieval settings}. The problem formulation of content-based document retrieval is defined as follows: In the first phase, which corresponds to the indexing phase, we extract the vision and language backbones and generate the embeddings for all document images -in the dataset in which GlobalDoc has been already pre-trained on- using the target modality only. In the second phase, which corresponds to the retrieval phase, we process the query modality using the pre-trained GlobalDoc model without activating (\ie with frozen backbones) the network of the target modality (\ie which can be either vision or language), as illustrated in~\cref{fig:document_retrieval}.
As a performance measure of the ranking of the retrieved results, we use the standard evaluation metric in content-based retrieval (\ie Recall@K (R@K)). 

\section{Experiments and Results}
\label{sec:results}
\begin{table*}[t!]
\centering
\begin{center}
    \resizebox{\linewidth}{!} {%
    \begin{tabular}{l|ccc|cc|cc|ccc|ccc}
    \hline
    Setting & \multicolumn{3}{c}{Pre-train Task} & \multicolumn{2}{c}{Meta-Learning} & \multicolumn{2}{c}{Head} & \multicolumn{6}{c}{Modality} \\
    & L2M & L2U & L2R & Meta-train & Meta-test & MLP & CMAE & \multicolumn{3}{c}{Vision} & \multicolumn{3}{c}{Language}\\
    
    \hline
    
    & & & & & & & & \multicolumn{3}{c}{\textbf{5-way/15-Query}} & \multicolumn{3}{c}{\textbf{5-way/15-Query}} \\
    & & & & & & & & \textbf{1-shot} & \textbf{5-shot} & \textbf{20-shot} & \textbf{1-shot} & \textbf{5-shot} & \textbf{20-shot} \\
    \cline{9-14}
    
    \multirow{4}{*}{S1} 
    & \cmark & \xmark & \xmark
    & \xmark & \cmark
    & \xmark & \xmark 
    & 34.66 $\pm$ 0.66 & 44.70 $\pm$ 0.64 & 51.15 $\pm$ 0.66  
    & 32.49 $\pm$ 0.63 & 41.73 $\pm$ 0.58 & 48.84 $\pm$ 0.54 \\

    & \cmark & \xmark & \xmark
    & \xmark & \cmark
    & \cmark & \xmark 
    & 41.33 $\pm$ 0.71 & 61.55 $\pm$ 0.65 & 75.05 $\pm$ 0.49
    & 38.02 $\pm$ 0.68 & 54.87 $\pm$ 0.66 & 68.92 $\pm$ 0.55 \\
    
    & \cmark & \xmark & \xmark
    & \cmark &\cmark
    & \xmark &\xmark 
    & 43.81 $\pm$ 0.71 & 63.63 $\pm$ 0.63 & 76.46 $\pm$ 0.51 
    & 38.91 $\pm$ 0.63 & 57.40 $\pm$ 0.61 & 72.57 $\pm$ 0.54 \\

    & \cmark & \xmark & \xmark
    & \cmark &\cmark
    & \cmark &\xmark & 
    \textbf{53.51} $\pm$ \textbf{0.80} & \textbf{74.48} $\pm$ \textbf{0.67} & \textbf{82.86} $\pm$ \textbf{0.51}  
    & \textbf{38.14} $\pm$ \textbf{0.61} & \textbf{57.04} $\pm$ \textbf{0.61} & \textbf{72.20} $\pm$ \textbf{0.56} \\
    
    \hline
    
    \multirow{2}{*}{S2} 
    & \cmark & \cmark & \xmark
    & \xmark & \cmark
    & \xmark & \cmark & 54.89 $\pm$ 0.83 & 74.58 $\pm$ 0.62 & 82.91 $\pm$ 0.49
    & 67.23 $\pm$ 0.96 & 77.82 $\pm$ 0.41 & 78.87 $\pm$ 0.35 \\

    & \cmark & \cmark & \xmark
    & \cmark & \cmark
    & \xmark & \cmark & \textbf{67.23} $\pm$ \textbf{0.96} & \textbf{77.82} $\pm$ \textbf{0.41} & \textbf{78.87} $\pm$ \textbf{0.35}
    & \textbf{67.01} $\pm$ \textbf{0.94} & \textbf{77.53} $\pm$ \textbf{0.42} & \textbf{78.86} $\pm$ \textbf{0.35} \\
    
    \hline
    
    \multirow{2}{*}{S3} 
    & \cmark & \cmark & \cmark
    & \xmark & \cmark
    & \xmark & \cmark & 79.08 $\pm$ 0.88 & 89.10 $\pm$ 0.39 & 89.96 $\pm$ 0.37
    & 75.45 $\pm$ 0.94 & 86.79 $\pm$ 0.41 & 88.45 $\pm$ 0.38 \\

    & \cmark & \cmark & \cmark
    & \cmark & \cmark
    & \xmark & \cmark  & \textbf{80.63} $\pm$ \textbf{0.64} & \textbf{89.36} $\pm$ \textbf{0.49} & \textbf{90.34} $\pm$ \textbf{0.38}
    & \textbf{79.77} $\pm$ \textbf{0.61} & \textbf{89.54} $\pm$ \textbf{0.56} & \textbf{90.33} $\pm$ \textbf{0.38} \\
    
   \hline
   
    \end{tabular}%
    }
\end{center}
\caption{Ablation study on the Few-shot Image Classification task on the RVL-CDIP dataset. All accuracy results are averaged over $600$ test episodes and reported with $95\%$ confidence intervals.}
\label{tab:table_5.2}
\end{table*}  
\begin{table*}[t]
\centering
\begin{center}
    \resizebox{\linewidth}{!} {%
        \begin{tabular}{l|ccc|ccc|ccc|ccc|ccc}
        \hline
        Setting & \multicolumn{3}{c}{Pre-train Task} & \multicolumn{6}{c}{Uni-Modal Retrieval} & \multicolumn{6}{c}{Cross-Modal Retrieval} \\
         & L2M & L2U & L2R & \multicolumn{3}{c}{Vision $\to$ Vision} & \multicolumn{3}{c}{Language $\to$ Language} &  \multicolumn{3}{c}{Vision $\to$ Language} &  \multicolumn{3}{c}{Language $\to$ Vision}\\
        
        \hline
        
        & & & & \textbf{R@1} & \textbf{R@5} & \textbf{R@10} & \textbf{R@1} & \textbf{R@5} & \textbf{R@10} & \textbf{R@1} & \textbf{R@5} & \textbf{R@10} & \textbf{R@1} & \textbf{R@5} & \textbf{R@10} \\

        \cline{5-16}
        
        S1
        & \cmark & \xmark & \xmark
        & 78.85 & 91.21 & 94.23 & 74.52 & 90.00 & 93.67
        & 5.37 & 14.60 & 21.20 & 4.73 & 13.30 & 19.29 \\
        \hline
        S2
        & \cmark & \cmark & \xmark
        & 80.63 & 92.06 & 94.83 & 75.15 & 90.39 & 93.96 
        & 73.05 & 89.48 & 93.41 & 70.11 & 86.05 & 91.34 \\
        \hline
        S3
        & \cmark & \cmark & \cmark
        & \textbf{82.85} & \textbf{93.15} & \textbf{95.49} & \textbf{79.00} & \textbf{92.07} & \textbf{95.03} 
        & \textbf{75.28} & \textbf{90.07} & \textbf{93.58} & \textbf{73.74} & \textbf{88.00} & \textbf{92.29} \\

        \hline

    \end{tabular}%
    }
\end{center}
\caption{R@K Quantitative evaluation results of uni-modal and cross-modal content-based retrieval on the RVL-CDIP test set.}
\label{tab:table_5.4}
\end{table*}   
\begin{table}[t]
\centering
\resizebox{\columnwidth}{!}{%
    \begin{tabular}{l|ccc}
    \hline
    Method & \multicolumn{3}{c}{Multi-Modal Modality} \\
    \hline
    & \multicolumn{3}{c}{\textbf{5-way/15-Query}} \\
    & \textbf{1-shot} & \textbf{5-shot} & \textbf{20-shot} \\
    \cline{2-4}
    LayoutLMv3$_{\text{Base}}$~\cite{huang2022layoutlmv3}
    & 26.92 $\pm$ 0.56 & 33.08 $\pm$ 0.62 & 38.17 $\pm$ 0.57 \\
    \hline
    \textbf{GlobalDoc}  
    & \textbf{80.20} $\pm$ \textbf{0.23} & \textbf{89.47} $\pm$ \textbf{0.53} & \textbf{90.35} $\pm$ \textbf{0.39} \\
    \hline
    \end{tabular}%
}
\caption{A comparison with LayoutLMv3$_{\text{Base}}$ pre-trained model on the Few-Shot Document Image Classification task on the RVL-CDIP test set.}
\label{tab:table_5.3}
\end{table}
\subsection{Experimental Settings}
\noindent\textbf{Pre-training.} we conduct pre-training on a subset of the IIT-CDIP~\cite{lewis2006building} dataset containing 1.4M document images to learn multimodal representations. GlobalDoc is initialized from the pre-trained weights of the pre-trained vision and language backbones. For the multimodal attention encoder, the weights are randomly initialized. We pre-train GlobalDoc using AdamW~\cite{loshchilov2017decoupled} optimizer with a batch size of $128$ for $499,600$ steps. We use a weight decay of $10^{-2}$. The learning-rate is warmed-up to $10^{-4}$ in the first $10\%$ iterations, and decayed to $5\cdot10^{-5}$ following a linear decay schedule. The temperature parameter $\tau$ is set to $0.07$, and the size of the queue used for L2M is set to $65,536$. \\
\noindent\textbf{Fine-tuning.} 
\textcolor{black}{we conduct the few-shot DIC, content-based DIR and DIC experiments on the RVL-CDIP dataset~\cite{harley2015evaluation}. It consists of gray-scale labeled documents split into 16 categories: Advertisement, Budget, Email, File folder, Form, Handwritten, Invoice, Letter, Memo, News article, Presentation, Questionnaire, Resume, Scientific publication, Scientific report, and Specification. The dataset is split into 320K training documents, 40K documents documents for validation and test sets. Besides, the task of visual information extraction (VIE) is conducted on the FUNSD dataset~\cite{jaume2019funsd}. This dataset comprises 199 noisy scanned documents, sampled from the RVL-CDIP dataset, annotated with labels for 9,707 semantic entities. The dataset is split into 149 training samples and 50 test samples.} 
\subsection{Fine-Tuning on Cross-Modal Tasks}
\subsubsection{Task I: Proposed Few-Shot DIC}
\noindent\textbf{How effective are the designed pretext objectives?}
As detailed in ~\cref{tab:table_5.2}, we perform an ablation study on the few-shot image classification task under three settings S1, S2 and S3. Each setting is performed w/wo our proposed pretext objectives (\ie L2M, L2M+L2U and L2M+L2U+L2R), w/wo meta-training, w/wo the MLP, and w/ the CMAE. For each setting, the best-performing method is highlighted in bold. The results are averaged over 600 experiments as in~\cite{chen2019closer, ravi2016optimization}. 
We conduct experiments on the most common setting in the conventional few-shot image classification task: 1-shot, 5-shot, and 20-shot. We use the pre-trained GlobalDoc as the embedding network, and perform 5-way classification for only novel classes. During meta-training, we split document samples onto 9 classes for fine-tuning, and 7 classes for testing. Note that we sample only 600 samples from each class as in~\cite{dhillon2019baseline}. The results show that the two-step pre-training approach (\ie S3) improves semantic representation learning, and thus improves the overall results of both uni-modal vision and language modalities compared to a one-step only pre-training approach (\ie S1 and S2). Also, with the use of the CMAE, the model learns better information by unifying the vision-language sample pairs. Furthermore, we observe that the performance of our proposed method significantly increases when receiving more samples as input (\ie 20-shot) w/wo meta-training. 

\noindent\textbf{Does the amount of pre-training data matters?} In this part, we conduct experiments on the multi-modal setting given multi-modal inputs in both offline and online stages to make the results a fair comparison with the LayoutLMv3$_{Base}$~\cite{huang2022layoutlmv3} baseline. We built our implementation using the released pre-trained model and performed inference (\ie online stage) on the few-shot document image classification task. The results in ~\cref{tab:table_5.2} illustrate that GlobalDoc outperforms the open-sourced available LayoutLMv3 model by a huge margin on the 1-shot, 5-shot and 20-shot classification settings. This shows that GlobalDoc is able to fulfill the \textit{objective of a decent transferable VDU model for an industrial online setting breaking the myth of pre-training millions of document samples}.  
\subsubsection{Task II: Proposed Content-based DIR}
\begin{figure*}[t!]
\centering
\begin{subfigure}{.24\textwidth}
  {
  \includegraphics[width=\linewidth]{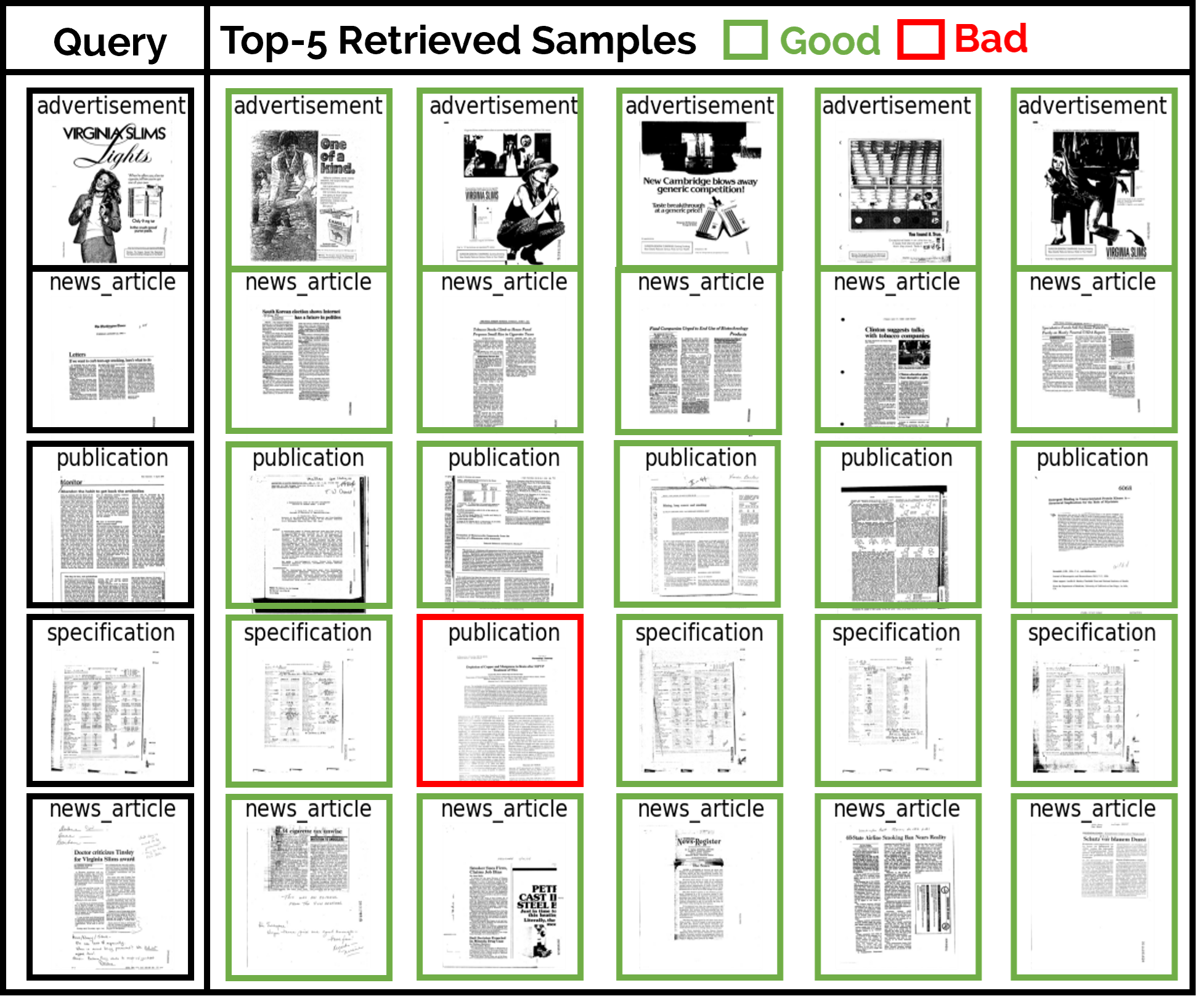}}\quad
  \caption{Vision to vision}
  \label{fig:vision_to_vision_retrieval}
\end{subfigure}
\begin{subfigure}{.24\textwidth}
  {
  \includegraphics[width=\linewidth]{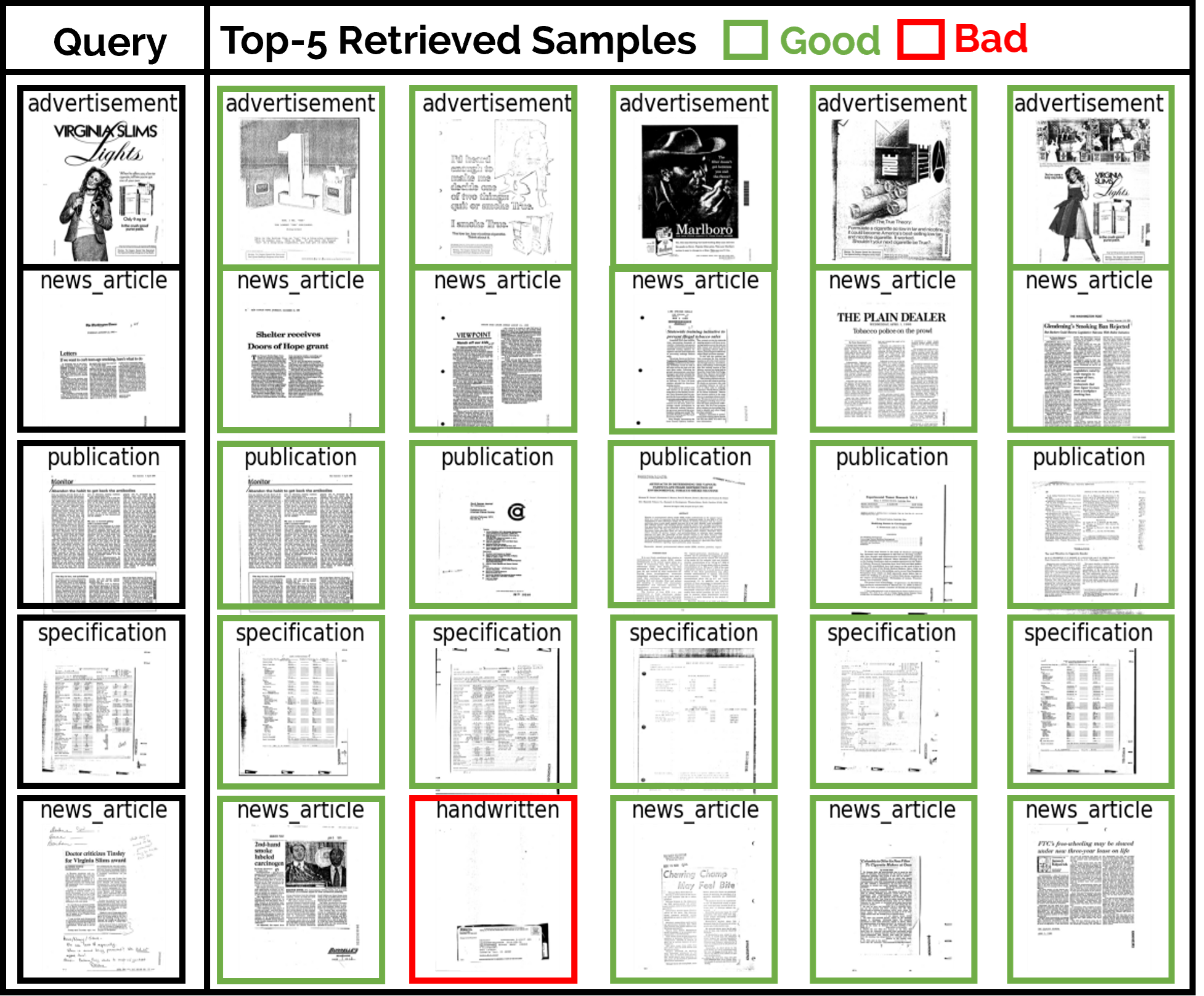}}\quad
  \caption{Lang to language}
  \label{fig:language_to_language_retrieval}
\end{subfigure}
\begin{subfigure}{.24\textwidth}
  {
  \includegraphics[width=\linewidth]{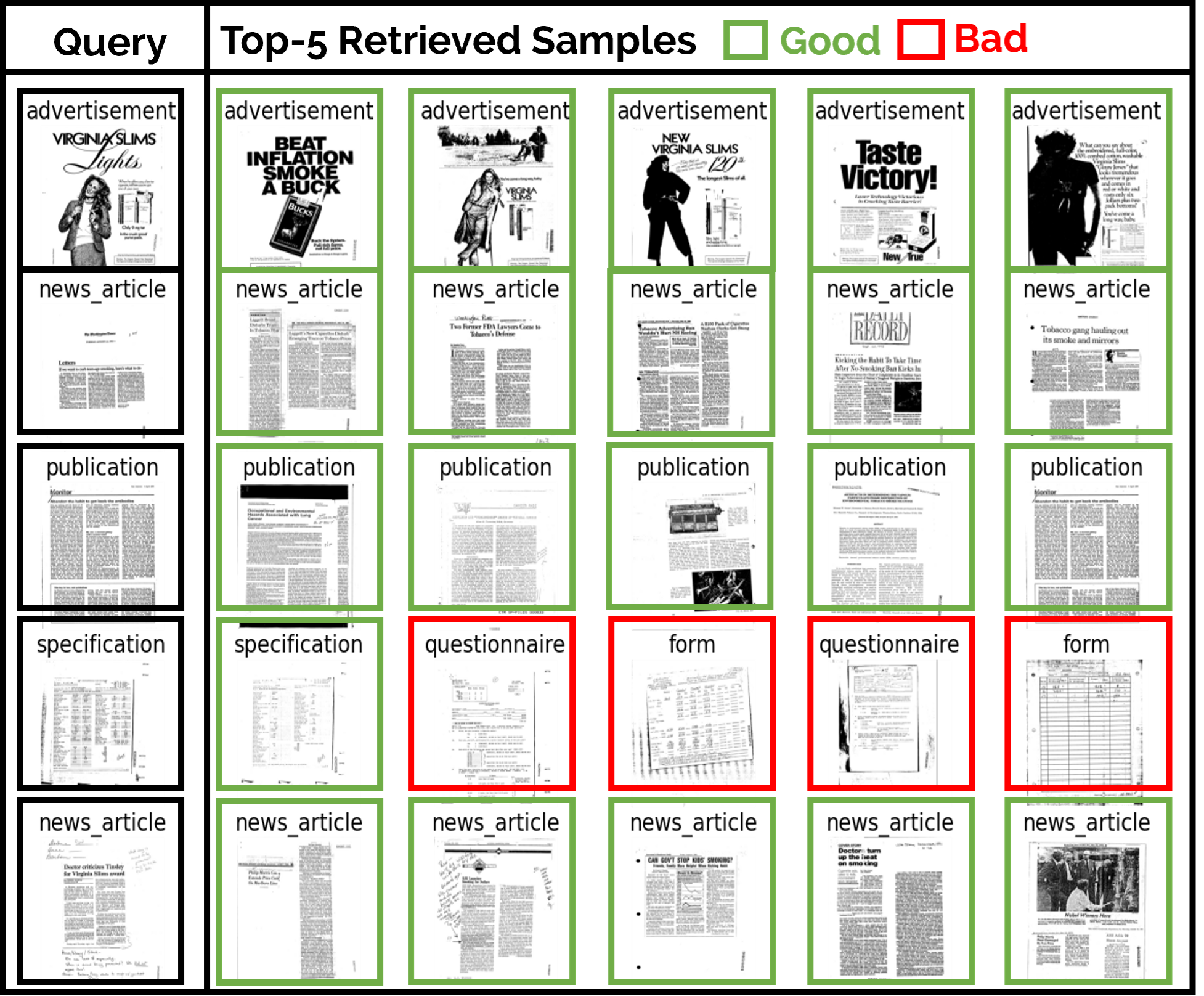}}\quad
  \caption{Vision to language}
  \label{fig:vision_to_language_retrieval}
\end{subfigure}
\begin{subfigure}{.24\textwidth}
  {
  \includegraphics[width=\linewidth]{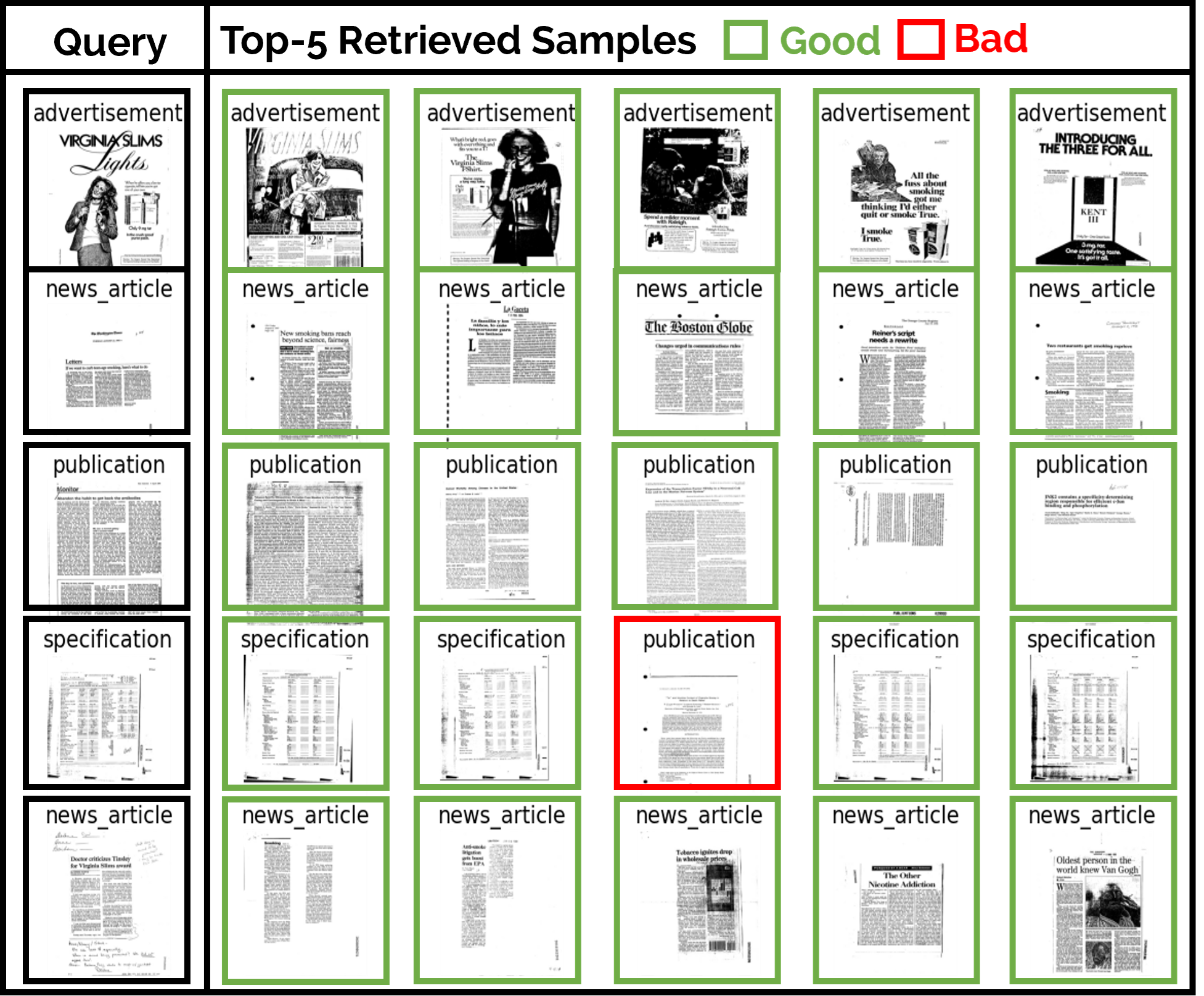}}\quad
  \caption{Language to vision}
  \label{fig:language_to_vision_retrieval}
\end{subfigure}
\caption{\textbf{Representative uni-modal and cross-modal retrieval samples of GlobalDoc.} Zoom in for better visualization. The first column represents the input query -randomly selected- from the test set of RVL-CDIP. The top-5 retrievals are shown in following columns in order. Red and green borders are used to depict the incorrect and correct classes of retrieved documents respectively. For each task setting, the same input example query is used.}
\label{fig:figure_3.3}
\end{figure*}

\noindent\textbf{How good is GlobalDoc in terms of model efficiency and flexibility?} As reported in ~\cref{tab:table_5.4}, we conduct experiments on content-based document retrieval in different settings S1, S2 and S3. We observe that in the uni-modal retrieval scenario for setting S1, GlobalDoc achieves decent performance in retrieving top-K relevant information which belongs to the same document category as the input query. 
However, in cross-modal retrieval, the performance drops R@K score significantly for both visn $\to$ lang and lang $\to$ visn tasks. This drop of performance is mainly due to the fact that at a high level, GlobalDoc did not identify any agreement between vision-language sample pairs (\eg the font size or character style for an image region is not confirmed by the semantic meaning of the corresponding language features). This means that those features were not amplified. This brings us to add the CMAE module in setting S2 to better capture the cross-modal interactions between vision and language modalities at a higher level with a unified cross-modal learning objective (\ie L2U), with the aim to overcome the problem of the first L2M pretext task when performed on cross-modal retrieval. As seen in ~\cref{tab:table_5.4} that unifying vision-language sample pairs do not only improve the uni-modal results, but also cross-modal ones with nearly $71.59\%$ for the image modality, and nearly $70.06\%$ for the text modality, given all R@K scores. 
This justifies the importance of unifying high-level visual and textual features in a cross-modal fashion. Lastly, with the last setting S3, we aim to learn richer semantic concepts by adding the L2R on top to improve the representation learning for vision and language modalities. As illustrated in ~\cref{tab:table_5.4}, the best R@K scores have been achieved with L2M+L2U+L2R combined for uni-modal and cross-modal retrieval tasks. Moreover, we conduct experiments on the multi-modal setting for a fair comparison with the LayoutLMv3$_{Base}$~\cite{huang2022layoutlmv3} baseline, given multi-modal inputs as query. The results in ~\cref{tab:table_5.5} illustrate that GlobalDoc outperforms the LayoutLMv3 model by a massive margin on the top-k retrieval scores. It is important to notice that the results that have been achieved are quite similar in all configurations (\ie uni-modal, cross-modal and multi-modal). Hence, \textit{GlobalDoc enables to produce more transferable and semantically rich embeddings}. We further show some qualitative samples of different retrieval scenarios given a challenging example from the "specification" and "report" categories as the input query in ~\cref{fig:figure_3.3}. We observe both good retrieval cases and some failure ones for both uni-modal and cross-modal retrieval settings. The layout of a "specification" document includes big structured tables which are also found in other classes like "form" or "questionnaire" or "publication". As well, the "report" category includes mostly text-only information having "memo" and "budget" document-like structures defined as the problem of inter-class variability which is challenging in the case of cross-modal retrieval scenarios, where there are more failure cases. However, we depict some interesting retrieval cases as in "advertisement", "news$\_$article", "publication", and "resume" queries.
\begin{table}[t]
\centering
    \begin{tabular}{l|ccc}
    \hline
     Method & \multicolumn{3}{c}{Multi-modal Modality} \\
   \hline
    & \textbf{R@1} & \textbf{R@5} & \textbf{R@10} \\
    \cline{2-4}
    LayoutLMv3$_{\text{Base}}$~\cite{huang2022layoutlmv3} & 32.62 & 60.29 & 73.95 \\
    \hline
    \textbf{GlobalDoc} & \textbf{80.93} & \textbf{92.62} & \textbf{95.27} \\
    \hline
    \end{tabular}%
\caption{Quantitative evaluation results of multi-modal Content-based DIR in terms of R@K on the RVL-CDIP test set.}
\label{tab:table_5.5}
\end{table}
\subsubsection{Task III: Standard Benchmark VDU Tasks}
\noindent\textbf{Document Image Classification (DIC).}
The document image classification task aims to predict the category of visually rich document images. The fine-tuning process takes $249,900$ iterations with a batch size of $64$ and a learning rate of $5\cdot10^{-5}$. We report in ~\cref{tab:table_5.1} the classification performance on the test set of the RVL-CDIP dataset, where the metric used is the top-1 accuracy. GlobalDoc achieves SOTA performance in uni-modal setting with 92.58\% accuracy (\textcolor{green}{+0.47}), outperforming DiT$_{Base}$, pre-trained on 42M document images. Also, under the uni-modal Text setting, GlobalDoc outperforms the baselines with a 93.82\% accuracy (\textcolor{green}{+2.04}) compared to Bert$_{Base}$ and RoBERTa$_{Base}$ and comparable results compared to LiLT$_{Base}$, BROS$_{Base}$ which rely on OCR token-level (\textcolor{red}{-1.86}). Moreover, under the (Vision+Text) setting, GlobalDoc achieves an accuracy of 94.04\%. Thus, GlobalDoc enables to reduce the gap (\textcolor{red}{-2.13}) with related VDU works pre-trained on extensive amount of pre-training document data under the (Vision+Text+Layout) setting in both offline and online stages.

\noindent\textbf{Visual Information Extraction (VIE).}
The goal of this task is to assign labels such as "question," "answer," "header," or "other" to each semantic entity. The officially provided OCRed text without the relative positional encodings (\ie layout information). We take the semantic entities as input and feed the final output representations to a classifier. We use the entity-level F1 score as the evaluation metric. The model is fine-tuned with a learning rate of $3\cdot10^{-5}$ and a batch size of 16 for 50 epochs. While the VIE task heavily rely on OCR to extract relative local positional encodings at the token-level, our approach is designed to leverage page-level information without relying on OCR. It is worth noting, as shown in~\cref{tab:table_5.1}, that SOTA VDU models under the setting (Vision+Text+Layout) like LayoutLM$_{v1,v2,v3}$, DocFormer, etc. achieve the best performance by integrating both token-local and/or region-level information. However, GlobalDoc, which relies on page-level information, still achieves comparable performance under the modality settings of (Text and Text+Layout), despite a gap in performance under the (Vision+Text+Layout) setting.

\begin{table}[t]
\centering
\begin{center}
\resizebox{\columnwidth}{!} {%
    {\begin{tabular}{@{}lrcc@{}}
    \hline
        \multirow{2}{*}{Method} & \multirow{2}{*}{\#Pretrain Data} & \multicolumn{1}{c}{RVL-CDIP} & \multicolumn{1}{c}{FUNSD} \\
        \multicolumn{1}{c}{} & \multicolumn{1}{c}{} & \multicolumn{1}{c}{Accuracy (\%)} & \multicolumn{1}{c}{F1 (\%)} \\
    \hline
    \textit{Vision-only methods} \\
    \hline
        \textbf{GlobalDoc} (V)   & 320k & 92.04 & -    \\
        DiT$_{Base}$~\cite{li2022dit}                       & 42M & 92.11 & -  \\
        \textbf{GlobalDoc} (V)   & 1.4M & \textbf{92.58} & -    \\

    \hline
    \textit{Text-only / (Text + Layout) methods}\\
    \hline
        BERT$_{Base}$~\cite{devlin2018bert}                 & - & 89.81 & 60.26     \\
        RoBERTa$_{Base}$~\cite{liu2019roberta}              & - & 90.06 & 66.48     \\
        LayoutLMv1$_{Base}$~\cite{xu2020layoutlm}             & 11M & 91.78 & 78.66 \\
        \textbf{GlobalDoc} (T)   & 320k & 93.13 & 62.27  \\
        \textbf{GlobalDoc} (T)   & 1.4M & 93.82 & 65.52  \\
        LiLT$_{Base}$~\cite{wang2022lilt}                   & 11M & \textbf{95.68} & 88.41 \\
        FormNet~\cite{lee2022formnet}                         & 700k & - & 84.69  \\
        BROS~\cite{hong2022bros}                            &  11M  & \underline{95.58} & 81.21 \\ 
        DocFormer$_{Base}$~\cite{appalaraju2021docformer}   & 5M & - & 80.54     \\
    \hline
    \textit{Vision + Text methods}\\
    \hline
        VLCDoc~\cite{bakkali2023vlcdoc}   & 320k  & 93.19 & - \\
    \textbf{GlobalDoc} (V+T)   & 320k & 93.18 & 77.84  \\
    \textbf{GlobalDoc} (V+T)   & 1.4M & \textbf{94.04} & 79.40  \\
    \hline
    \textit{Vision + Text + Layout methods}\\
    \hline
        SelfDoc~\cite{li2021selfdoc}                        & 320k & 92.81 & 83.36         \\
        LayoutLMv1$_{Base}$~\cite{xu2020layoutlm}             & 11M & 94.42 & 79.27     \\
        UDoc~\cite{gu2022unified}                           & 11M & 95.05 & 87.93     \\
        TILT$_{Base}$~\cite{powalski2021going}              & 1M & 95.25 & -     \\
        LayoutLMv2$_{Base}$~\cite{xu2020layoutlmv2}         & 11M & 95.25 & 82.76     \\
        LayoutLMv3$_{Base}$~\cite{huang2022layoutlmv3}      & 11M & 95.44 & \textbf{90.29}       \\
        DocFormer$_{Base}$~\cite{appalaraju2021docformer}   & 5M & \textbf{96.17} & 83.34     \\
        GraphDoc$_{Resnet}$~\cite{zhang2022multimodal}  & 320k & 96.10 & 87.95     \\
        GraphDoc~\cite{zhang2022multimodal}             & 320k & 96.02 & 87.77      \\
    \hline
    \end{tabular}}%
    }
\end{center}
\caption{RVL-CDIP and FUNSD comparison results of different VDU methods pre-trained in an SSL-based fashion. V, T, and L denote Vision, Text, and Layout modalities. \textbf{Bold} indicates the SOTA, and \underline{underline} indicates the second best.}
\label{tab:table_5.1}
\end{table}

\section{Conclusion}
\label{sec:conclusion}
In this work, we approach the document representation learning problem by addressing the issue of VDU models relying on 2D word position embeddings and the extensive amount of pre-training document collections. We explore an industrial online setting where new classes appear regularly and large annotated dataset can not be used. To deal with these constraints, we proposed a more generalizable and flexible VDU model named GlobalDoc which effectively incorporate cross-modal representations from language and vision for document-level downstream applications. It exhibits compelling performance over SOTA models in both uni-modal and cross-modal tasks. Further research in this representation learning framework could be highly beneficial for the DocumentAI community.

{\small
\bibliographystyle{ieee_fullname}
\bibliography{egbib}
}

\end{document}